\PassOptionsToPackage{table,dvipsnames}{xcolor}

\documentclass{maskwam}

\usepackage{amsmath}
\usepackage{amssymb}
\usepackage{amsfonts}
\usepackage{mathtools}
\usepackage{enumitem}
\usepackage{algorithm}
\usepackage{algorithmic}
\usepackage{wrapfig}
\usepackage{adjustbox}
\usepackage{tabularx}
\usepackage{makecell}
\usepackage{siunitx}
\usepackage{float}
\usepackage{diagbox}
\usepackage{dsfont}

\definecolor{bestgreen}{RGB}{224,232,240}
\definecolor{secondgreen}{RGB}{240,245,249}

\title{MaskWAM: Unifying Mask Prompting and Prediction for World-Action Models}

\renewcommand{\authorlist}{%
  {\sffamily \sbseries
  Hanyang Yu      $^{1,\ddagger}$,
  Haitao Lin      $^{2,\dagger}$,
  Jingbo Zhang    $^{2}$,
  Wenyao Zhang    $^{2}$,\\
  Chenghao Gu     $^{4,\ddagger}$,
  Heng Li         $^{1}$,
  Ping Tan        $^{1,\dagger}$
  }%
}

\affiliation[1]{The Hong Kong University of Science and Technology}
\affiliation[2]{Tencent Robotics X}
\affiliation[3]{Tsinghua University}

\contribution[\ddagger]{Work done during an internship at Tencent Robotics X}
\contribution[\dagger]{Corresponding author}

\abstract{
World Action Models (WAMs) present a promising paradigm for robotic control via video prediction. However, current WAMs suffer from fundamental spatial bottlenecks: standard text inputs introduce referential ambiguity in cluttered scenes, while unstructured RGB predictions lack semantic grounding and remain biased by task-irrelevant backgrounds. To overcome these limitations, we introduce MaskWAM, an object-centric world-action model. By jointly integrating masks as both explicit inputs and predictions via a unified Mixture of Transformers (MoT), MaskWAM unlocks robust policy generalization. This design provides two key benefits: (1) predicting future masks yields object-centric semantic supervision that suppresses visual noise, significantly enhancing even standard text-conditioned WAMs; and (2) coupling this predictive supervision with first-frame visual prompts, such as target object masks, establishes a precise spatial anchor that substantially reduces language ambiguity. Crucially, as WAMs are inherently vision-driven architectures, direct mask conditioning yields substantially stronger guidance than text alone, establishing a precise and robust paradigm for manipulating unseen objects. Evaluations on LIBERO, RoboTwin, and real-world tasks demonstrate that MaskWAM significantly outperforms baselines in both language-clear and language-ambiguous tasks.
}

\date{\sffamily\today}

\metadata[Project Page]{\url{https://hanyangyu1021.github.io/maskwam.github.io/}}
\metadata[Keywords]{World-Action Models, Mask Visual Prompting, Object-Centric Manipulation}

\begin{document}

\maketitle

\section{Introduction}
\label{sec:intro}

World Action Models (WAMs)~\cite{bi2025motus, yuan2026fast,li2026causal,kim2026cosmos,pai2025mimic,zhang2026disentangled} have emerged as a promising alternative to Vision-Language-Action (VLA) policies~\cite{black2024pi0,black2025pi0.5,lin2026universal,team2026hy,wu2026pragmatic,liu2025rdt,dreamvla25,sun2026vla} by coupling action generation with future video prediction. Rather than directly mapping observations and language to actions, WAMs treat the modeling of future observations as a powerful proxy task~\cite{yuan2026fast}. This paradigm is highly appealing because it fundamentally enhances the underlying representation of the model. By capturing physical dynamics and task-relevant temporal structures, these predictive representations more effectively utilize spatio-temporal priors, thereby facilitating better transfer to downstream control tasks.

\begin{figure}[htb]
\centering
\includegraphics[width=1.0\textwidth]{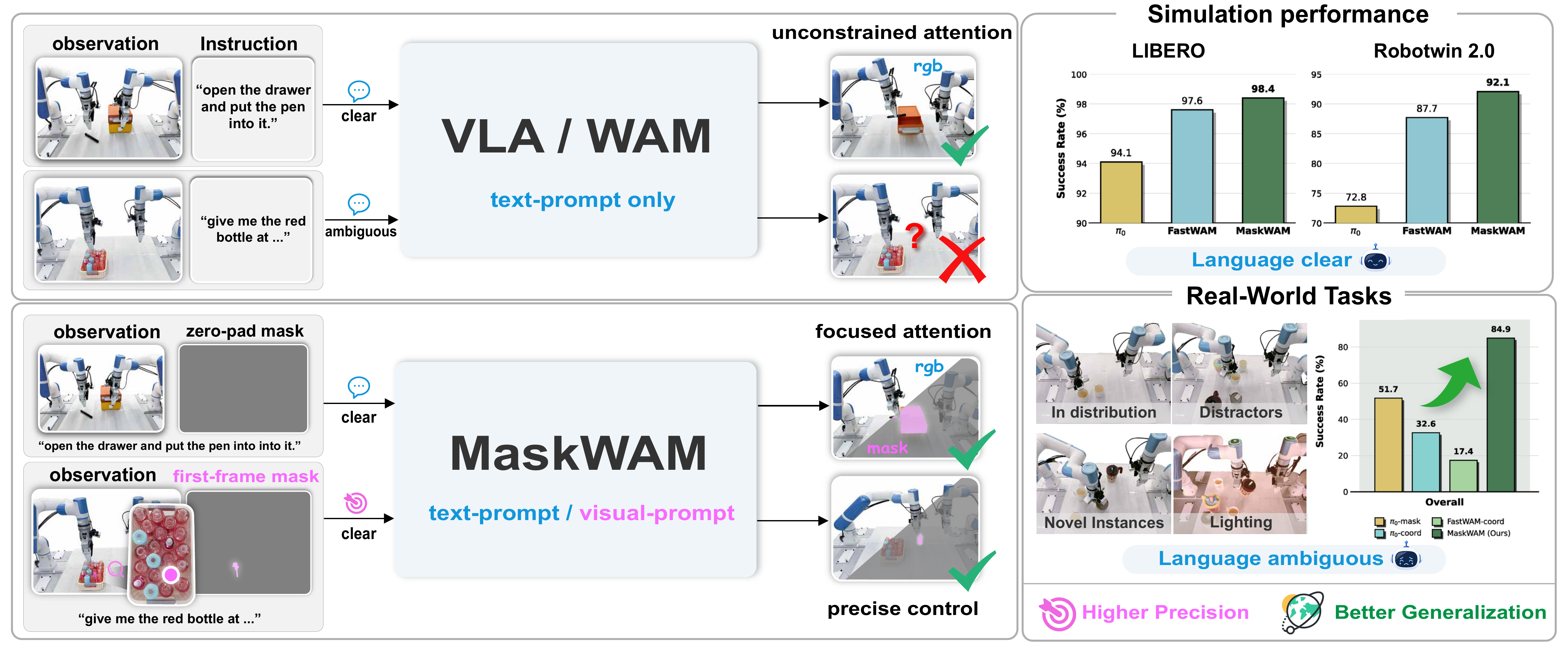}
\vspace{-4mm}
\caption{\textbf{Overview.} We introduce MaskWAM, an end-to-end WAM that unifies mask prompting and mask prediction, supporting both text prompts and visual prompts. MaskWAM achieves high-precision manipulation and strong generalization in both simulation and real-world tasks.}
\label{fig:overview}
\vspace{-4mm}
\end{figure}

However, deploying WAMs in real-world tasks presents two critical challenges in target representation, specifically regarding how the model encodes the target object and its spatial location: (1) \textit{Lack of explicit grounding in pure visual predictions.} While forecasting RGB frames yields implicit spatial priors, visual reconstruction supervision remains weakly structured, treating all regions equally rather than prioritizing task-relevant objects. Consequently, the target can remain entangled with background clutter, degrading policy performance. To achieve reliable target isolation, the policy therefore requires auxiliary, explicit semantic supervision. (2) \textit{Inherent descriptive limits of textual conditioning.} Although textual instructions provide essential semantic context, they lack sufficient precision for exact spatial anchoring. Language is inherently limited in describing complex spatial relationships or identifying the intended object among visually similar distractors, as illustrated in Fig.~\ref{fig:overview}, leading to suboptimal policy guidance.

To overcome these spatial bottlenecks, we propose \textbf{MaskWAM}, an end-to-end world-action model that achieves precise control by unifying future mask prediction and visual prompting. This unified framework directly addresses the aforementioned challenges through two synergistic mechanisms. (1) \textit{Semantic supervision via mask prediction:} To tackle the lack of explicit grounding, MaskWAM predicts future masks alongside RGB frames. This provides the necessary highly semantic supervision signal to abstract away environmental noise and force the policy to inherently prioritize task-relevant regions. (2) \textit{Spatial anchoring via visual prompting:} To overcome the descriptive limits of textual conditioning, MaskWAM integrates first-frame visual prompts such as target masks. This establishes a precise spatial anchor that substantially reduces referential ambiguity, providing reliable guidance for the policy even in severely cluttered scenes.

Specifically, our architecture processes the current RGB alongside an \textit{optional} visual prompt anchored exclusively on the first frame of the episode. A frozen Video VAE~\cite{wan2025wan} compresses these inputs into latents, which are combined with noisy future latents, robot proprioception, and T5 language embeddings~\cite{raffel2020exploring}. This unified stream is processed by a Mixture of Transformers (MoT)~\cite{liang2024mixture} to jointly predict future RGB, masks, and actions via flow-matching. Crucially, explicitly predicting future masks prevents representational shortcuts, forcing the model to both actively attend to the initial visual anchor and prioritize task-relevant regions. Furthermore, since WAMs are inherently vision-driven, direct mask conditioning aligns naturally with their visual prediction process. Our experiments further show that this design provides more effective fine-grained guidance than the evaluated visual-prompted VLA baseline and text-conditioned WAMs or VLAs.

In summary, our main contributions are as follows: (1) We propose {MaskWAM}, a world-action model that integrates masks as explicit inputs and outputs, jointly modeling future RGB frames, task-relevant masks, and action chunks within a unified architecture. (2) We demonstrate the dual benefits of this design: predicting masks acts as a powerful semantic regularizer steering attention to target regions, while first-frame visual prompting provides a precise spatial anchor for target disambiguation, driving robust generalization across unseen objects. (3) Comprehensive evaluations show that MaskWAM achieves state-of-the-art success rates of 98.4\% on LIBERO and 92.2\% on RoboTwin. In real-world deployments, it attains 84.3\% on language-clear tasks and 84.9\% on challenging language-ambiguous scenarios, outperforming the strongest baseline by a 33.2\%.
\section{Related Work}

\noindent\textbf{Video World Models for Robotic Manipulation.}
Recent manipulation systems increasingly move beyond direct action regression toward models that explicitly predict future observations alongside future actions~\cite{du2023learninguniversalpoliciestextguided, wu2023unleashing, zhou2024robodreamer, feng2025vidarembodiedvideodiffusion, bharadhwaj2024gen2act, won2025dualstreamdiffusionworldmodelaugmented, cheang2024gr2generativevideolanguageactionmodel, jang2025dreamgenunlockinggeneralizationrobot, yu2024lm, zhao2025cotvlavisualchainofthoughtreasoning, cen2025rynnvla, cen2025worldvla, zhou2025act2goalworldmodelgeneral, zheng2025flarerobotlearningimplicit, dreamvla25}. Video-diffusion world models leverage large-scale generative priors to forecast plausible futures, providing richer supervision than behavioral cloning and a natural mechanism for reasoning about consequences before acting. However, most WAM formulations~\cite{zhu2025unifiedworldmodelscoupling, liang2025videogenerators, kim2026cosmos, ge2025, pai2025mimic, lingbot-va2026, bi2025motus, ye2026worldactionmodelszeroshot,hu2024video,ma2026dit4dit} still supervise the visual future mainly through RGB reconstruction. This leaves a representation gap between high-dimensional pixels and the lower-dimensional spatial structure that governs control. MaskWAM addresses this gap by adding future mask prediction for task-relevant regions, jointly trained with RGB futures and actions.

\noindent\textbf{Visual Prompting and Intermediate Representations for Robot Policies.}
To achieve robust spatial grounding, existing approaches either utilize visual cues as static inputs or predict them as intermediate representations. The former approach conditions policies on visual prompts (e.g., point clicks, bounding boxes, masks) to resolve referential ambiguity and explicitly isolate task-relevant objects~\cite{gu2023rttrajectory,bharadhwaj2024track2act,xu2024im2flow2act,sundaresan2023kite,fang2024moka,stone2023moo,huang2025roboground,yu2025point,li2025controlvla,liu2024moka}. Concurrently, the latter integrates intermediate representations like keypoints~\cite{huang2024rekep, yuan2024robopoint,haldar2025point,wang2025skil}, trajectories~\cite{hsu2025spot, lee2025molmoact, bharadhwaj2024track2act}, masks~\cite{mwm}, object flows~\cite{xu2024flow,wen2023any}, or affordance maps~\cite{ju2024robo,bahl2023affordances,nasiriany2025rt, nagarajan2019grounded} to bridge the gap between raw vision and continuous control. Unlike methods requiring per-frame conditional masks~\cite{li2025controlvla} or treating them as isolated targets~\cite{mwm}, MaskWAM unifies both paradigms. By combining first-frame visual conditioning with future mask prediction, it enforces explicit semantic supervision and steers attention toward the prompted target for robust spatial grounding.

\vspace{-1mm}
\section{Method}
\label{sec:method}
\begin{figure}[t]
\centering
\includegraphics[width=1.0\textwidth]{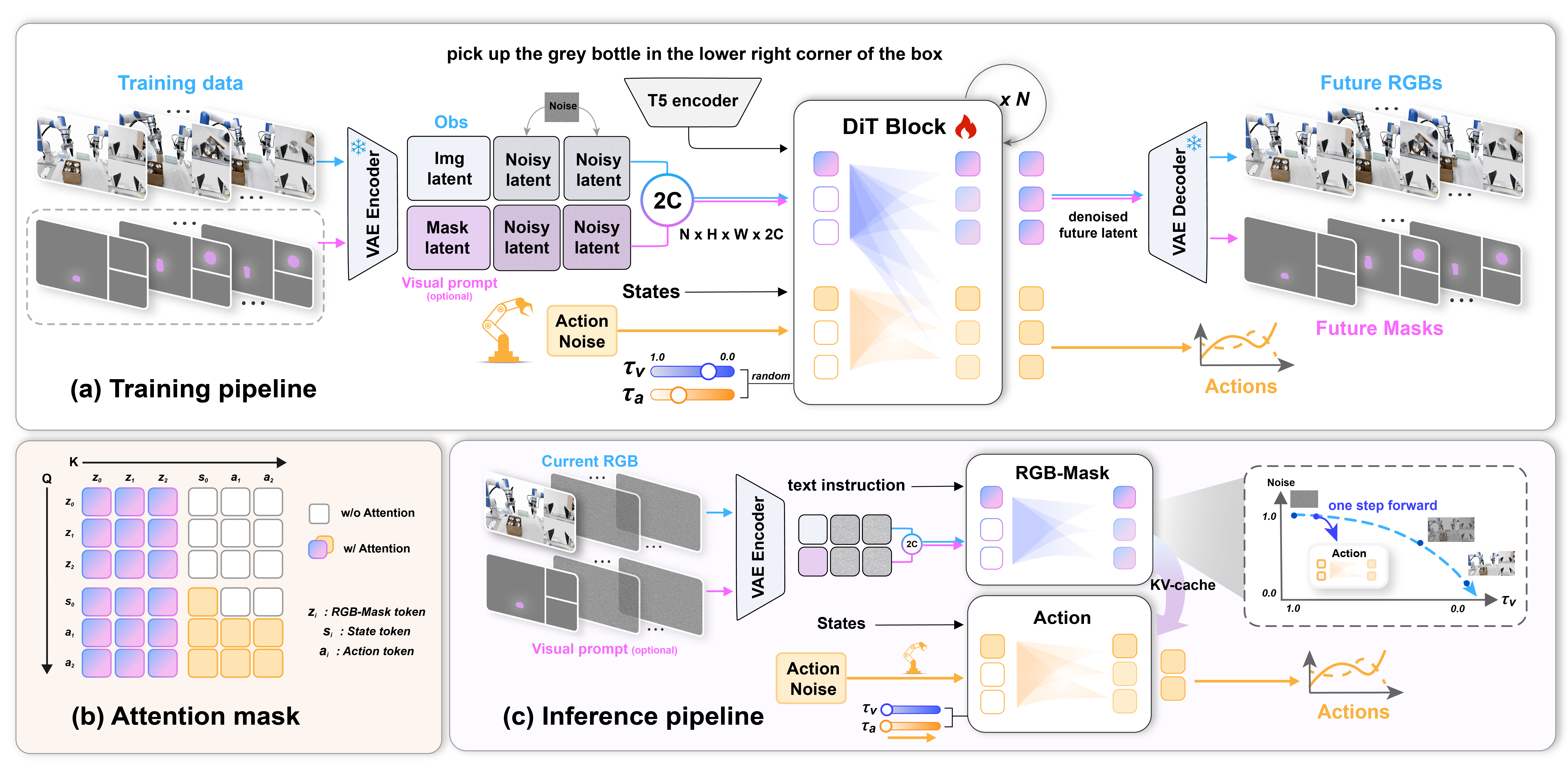}
\vspace{-6mm}
\caption{\textbf{MaskWAM architecture.} 
\textit{(a) Training:} Noisy RGB and mask latents are channel-concatenated and denoised by a unified DiT. The model jointly optimizes future RGB, future masks, and action chunking under a joint flow-matching framework with decoupled noise schedules $\tau_v$ and $\tau_a$. 
\textit{(b) Attention mask:} A block-wise causal attention mask enables unified RGB, mask, and action training.
\textit{(c) Inference:} Conditioned on optional first-frame masks to resolve spatial ambiguities, the model leverages KV-caching to efficiently generate actions from partially denoised latents.}
\label{fig:pipeline}
\vspace{-3mm}
\end{figure}

\vspace{-1mm}
\noindent\textbf{Overview.} Given a current RGB observation $\mathbf{I}_0 \in \mathbb{R}^{3 \times H \times W}$, a proprioceptive state $\mathbf{s}_0 \in \mathbb{R}^D$, a language instruction $\ell$, and an optional first-frame target mask $\mathbf{M}_0 \in \{0, 1\}^{H \times W}$, MaskWAM jointly predicts an action chunk $\mathbf{a}_{1:K} \in \mathbb{R}^{K \times D}$, future RGB frames $\mathbf{I}_{1:T} \in \mathbb{R}^{T \times 3 \times H \times W}$, and future masks $\mathbf{M}_{1:T}$. Formally, the model parameterizes the joint distribution
\begin{equation}
  p_\theta\bigl(\mathbf{a}_{1:K},\ \mathbf{I}_{1:T},\ \mathbf{M}_{1:T}\bigm| \mathbf{I}_0,\ \mathbf{M}_0,\ \mathbf{s}_0,\ \ell\bigr),
  \label{eq:problem}
\end{equation}
where $K$ is the action chunk size and $D$ is the action dimension. Built on Wan 2.2~\cite{wan2025wan}, MaskWAM extends conventional world-action models beyond RGB-space future prediction by introducing masks as explicit spatial representations of task-relevant regions. The action branch is implemented as a Mixture of Transformers (MoT) action expert, allowing video and action tokens to interact through joint attention as in Fig.~\ref{fig:pipeline}.

\subsection{Future Masks as Predictive Targets}
\label{sec:codiff}

\paragraph{Unified RGB and mask encoding.}
To use masks as prediction targets while preserving the visual priors of the pretrained video backbone, we represent each mask as an RGB-compatible image rather than introducing a separate mask-specific encoder. Specifically, a task-relevant mask is rendered as a three-channel image with a fixed color palette, while the background is assigned a separate color. The rendered mask sequence has the same temporal horizon and spatial resolution as the corresponding RGB observations. Both RGB frames and rendered mask frames are then encoded by the same causal 3D VAE encoder $\mathcal{E}$, producing RGB latents $\mathbf{z}_v \in \mathbb{R}^{C \times L \times H' \times W'}$ and mask latents $\mathbf{z}_m \in \mathbb{R}^{C \times L \times H' \times W'}$, where $C$, $L$, $H'$, and $W'$ denote the latent channel size, temporal length, spatial height, and spatial width, respectively.

The RGB and mask latents are concatenated along the channel dimension before being fed into the diffusion backbone:
\begin{equation}
  \mathbf{z} = [\mathbf{z}_v;\mathbf{z}_m] \in \mathbb{R}^{2C \times L \times H' \times W'}.
  \label{eq:concat}
\end{equation}
This design aligns mask information with the pretrained visual latent space and allows each latent position to encode both appearance and task-relevant spatial guidance. By treating masks as RGB-compatible inputs, MaskWAM can reuse the original VAE and transformer interface with minimal architectural changes, while still providing an explicit spatial representation for future prediction and action generation.

\paragraph{Patch embedding expansion.}
To process concatenated RGB and mask latents, we expand the input patch embedding of the pretrained video backbone from $C$ to $2C$ input channels. The original visual channels inherit the pretrained weights, while the added mask channels are initialized to zero. This preserves the initial behavior of the pretrained model and allows mask information to be gradually incorporated during fine-tuning. When mask prediction is enabled, the output head is also expanded from $C$ to $2C$ channels to predict both RGB and mask latent velocities.

\subsection{Initial Masks as Policy Conditions}
% \subsection{Visual-Prompt Controllability}
\label{sec:visual-prompt}

At deployment, $\mathbf{M}_0$ acts as an optional first-frame anchor that can be obtained from a text phrase, click, bounding box, or coarse mask via a segmentation model such as SAM-3~\cite{carion2025sam}. This anchor designates the target object or contact region in language-ambiguous scenes, such as duplicate objects or sub-object manipulation. MaskWAM does not require a prompt to benefit from future mask prediction, but the correspondence between the first-frame mask and the predicted mask future makes visual prompting a natural application: the prompt initializes the mask channel at the first frame, and the world model propagates that target specification into future RGB, future masks, and actions.

\paragraph{Unified Prompting via Mask Dropout.} 
To enable a single policy to seamlessly handle both language-clear tasks and language-ambiguous tasks, we employ an input mask dropout strategy during training. Specifically, the initial target mask $\mathbf{M}_0$ is replaced with a zero tensor with probability $p=0.5$. This unified training paradigm ensures perfect compatibility across different instruction regimes. Consequently, at deployment, the exact same model can dynamically incorporate an initial mask for precise target disambiguation, or rely exclusively on standard RGB and language inputs when the text instruction is inherently unambiguous.

\subsection{Training Objective: Joint Flow Matching}
\label{sec:loss}

Building upon the fused RGB and mask latent stream, MaskWAM utilizes a MoT architecture comprising two interacting branches: a visual branch that jointly denoises the visual and mask representations, and a lightweight action expert dedicated to denoising action chunks. Language features, extracted via a frozen T5 text encoder, are injected into the visual branch through cross attention. Concurrently, the proprioceptive state and noisy actions are processed through projection layers to serve as inputs for the action expert. This expert subsequently utilizes joint attention mechanisms over the visual context provided by the visual branch to predict action velocities.

Consequently, we train MaskWAM end to end using a unified flow matching objective that decouples the noise scheduling across two distinct domains. First, within the visual domain, the RGB and mask streams are synchronized under a shared visual timestep $\tau_v$. This ensures that visual appearance and structural masks remain spatiotemporally aligned throughout the generative process. Second, within the action domain, we sample an independent timestep $\tau_a$. This decoupling forces the action expert to perform trajectory denoising conditioned upon visual contexts at various noise levels. Crucially, this temporal independence establishes the theoretical foundation for the partial denoising inference strategy similar to~\cite{pai2025mimic}. The overall training objective is formulated as the sum of the individual flow matching losses:
\begin{equation}
  \mathcal{L} = \mathcal{L}_{\text{video}} + \mathcal{L}_{\text{mask}} + \mathcal{L}_{\text{act}},
  \label{eq:total-loss}
\end{equation}
where $L_{\mathrm{video}}$, $L_{\mathrm{mask}}$, and $L_{\mathrm{act}}$ denote the flow-matching losses for RGB videos, mask representations, and action trajectories, respectively.

\noindent\textbf{Implementation Details.}
% \label{sec:impl}
During training, the Video VAE and text encoder remain frozen. The architecture processes $384 \times 320$ RGB frames (T-shape) with a prediction horizon of $T=8$. 
% We use the AdamW optimizer, bfloat16 mixed precision, and FlashAttention 2. 
At deployment, the initial spatial anchor $\mathbf{M}_0$ is generated via SAM3~\cite{carion2025sam} \textit{only once at the start of the episode}, eliminating the need for real-time tracking and repeated visual prompting during subsequent predictions. To minimize latency, MaskWAM employs a partial-denoising decoding strategy. Rather than fully denoising future frames, the model performs only one denoising step on the joint RGB-mask stream to extract intermediate, task-aware visual latents. The action expert then jointly attends to these partially denoised latents, language instructions, and proprioception to generate the action chunk as in Fig.~\ref{fig:pipeline}(c). This avoids the high test-time cost of full video generation while preserving the mask-grounded world state. More details are provided in the appendix.

\vspace{-2mm}
\section{Experiments}

\paragraph{Benchmarks and Baselines.}
We evaluate MaskWAM on the LIBERO benchmark~\cite{liu2023libero} against leading baselines, including WorldVLA, GR00T-N1, $\pi_0$, $\pi_{0.5}$, Motus, and FastWAM, to show that auxiliary mask prediction improves policy learning even without test-time visual prompts. To assess multitask generalization, we evaluate MaskWAM on RoboTwin 2.0~\cite{chen2025robotwin} against $\pi_0$ and FastWAM under randomized instructions, environments, and object placements.
\vspace{-2mm}

\paragraph{Performance on LIBERO.}
As shown in Table~\ref{tab:libero}, MaskWAM establishes a new state-of-the-art with a 98.4\% average success rate. It consistently outperforms recent VLAs like $\pi_{0.5}$ and advanced WAMs such as Motus and FastWAM. Compared to our RGB-only variant, the auxiliary mask prediction objective lifts performance from 97.3\% to 98.4\%, showing that mask supervision can enhance base policy learning without visual prompts at deployment. Attention maps in the right panel of Table~\ref{tab:libero} explain this improvement: while the RGB-only model often highlights spurious backgrounds, mask supervision forces MaskWAM to attend precisely to task-relevant regions, yielding superior visual grounding and execution robustness.
\vspace{-2mm}

\paragraph{Performance on RoboTwin 2.0.}
As shown in Table~\ref{tab:robotwin_wide}, MaskWAM achieves a state-of-the-art 92.2\% average success rate across six randomized RoboTwin 2.0 tasks, outperforming $\pi_0$ and FastWAM by 19.4\% and 4.5\%, respectively. Ablations highlight the advantage of joint prediction: the Mask-only variant (88.8\%) natively outperforms RGB-only (87.3\%), but unifying both modalities maximizes performance to 92.2\%. This confirms that while RGB futures capture essential dynamics, auxiliary mask futures force the model to focus on task-relevant  regions. Notably, the good performance of our Mask-only ablation conceptually aligns with the findings of the concurrent Mask World Model [55], while our full MaskWAM further demonstrates that our joint representation achieves the best performance.
\vspace{-2mm}

\begin{table*}[t]
  \centering
  \scriptsize
  \setlength{\tabcolsep}{3.0pt}
  \renewcommand{\arraystretch}{1.08}
\caption{\textbf{Success rates (\%) on the LIBERO benchmark.} 
  The right panel visualizes action-to-video attention maps of RGB-based WAM and MaskWAM, showing that mask supervision encourages the model to attend more consistently to task-relevant regions. }
  \label{tab:libero}
  \vspace{-1mm}
  \begin{minipage}[t]{0.55\textwidth}
    \centering
    \vspace{0pt}
    \resizebox{\linewidth}{!}{
    \begin{tabular}{llccccc}
      \toprule
      \textbf{Method} & \textbf{Type} & \textbf{Spatial} & \textbf{Object} & \textbf{Goal} & \textbf{Long} & \textbf{Avg} \\
      \midrule
      WorldVLA~\cite{cen2025worldvla}        & VLA & 87.6 & 96.2 & 83.4 & 60.0 & 81.8 \\
      GR00T-N1~\cite{bjorck2025gr00t}        & VLA & 94.4 & 97.6 & 93.0 & 90.6 & 93.9 \\
      $\pi_0$~\cite{black2024pi0}            & VLA & 96.8 & 98.8 & 95.8 & 85.2 & 94.1 \\
      $\pi_{0.5}$~\cite{black2025pi0.5}      & VLA & \cellcolor{secondgreen}\underline{98.6} & 98.2 & \cellcolor{secondgreen}\underline{98.0} & 92.4 & 96.8 \\
      Motus~\cite{bi2025motus}               & WAM & 96.8 & \cellcolor{secondgreen}\underline{99.8} & 96.6 & \cellcolor{bestgreen}\textbf{97.6} & \cellcolor{secondgreen}\underline{97.7} \\
      FastWAM~\cite{yuan2026fast}            & WAM & 98.2 & \cellcolor{bestgreen}\textbf{100.0} & 97.0 & 95.2 & 97.6 \\
      \midrule
      Ours (RGB-only)                     & WAM & 96.8 & 99.6 & 97.0 & 95.8 & 97.3 \\
      Ours (Mask-only)                     & WAM & 97.2 & \cellcolor{secondgreen}\underline{99.8} & 97.4 & 96.0 & 97.6 \\
      Ours                      & WAM & \cellcolor{bestgreen}\textbf{98.8} & \cellcolor{bestgreen}\textbf{100.0} & \cellcolor{bestgreen}\textbf{98.2} & \cellcolor{secondgreen}\underline{96.4} & \cellcolor{bestgreen}\textbf{98.4} \\
      \bottomrule
    \end{tabular}
    }
  \end{minipage}
  \hfill
  \begin{minipage}[t]{0.4\textwidth}
    \centering
    \vspace{0pt}
    \includegraphics[width=\linewidth]{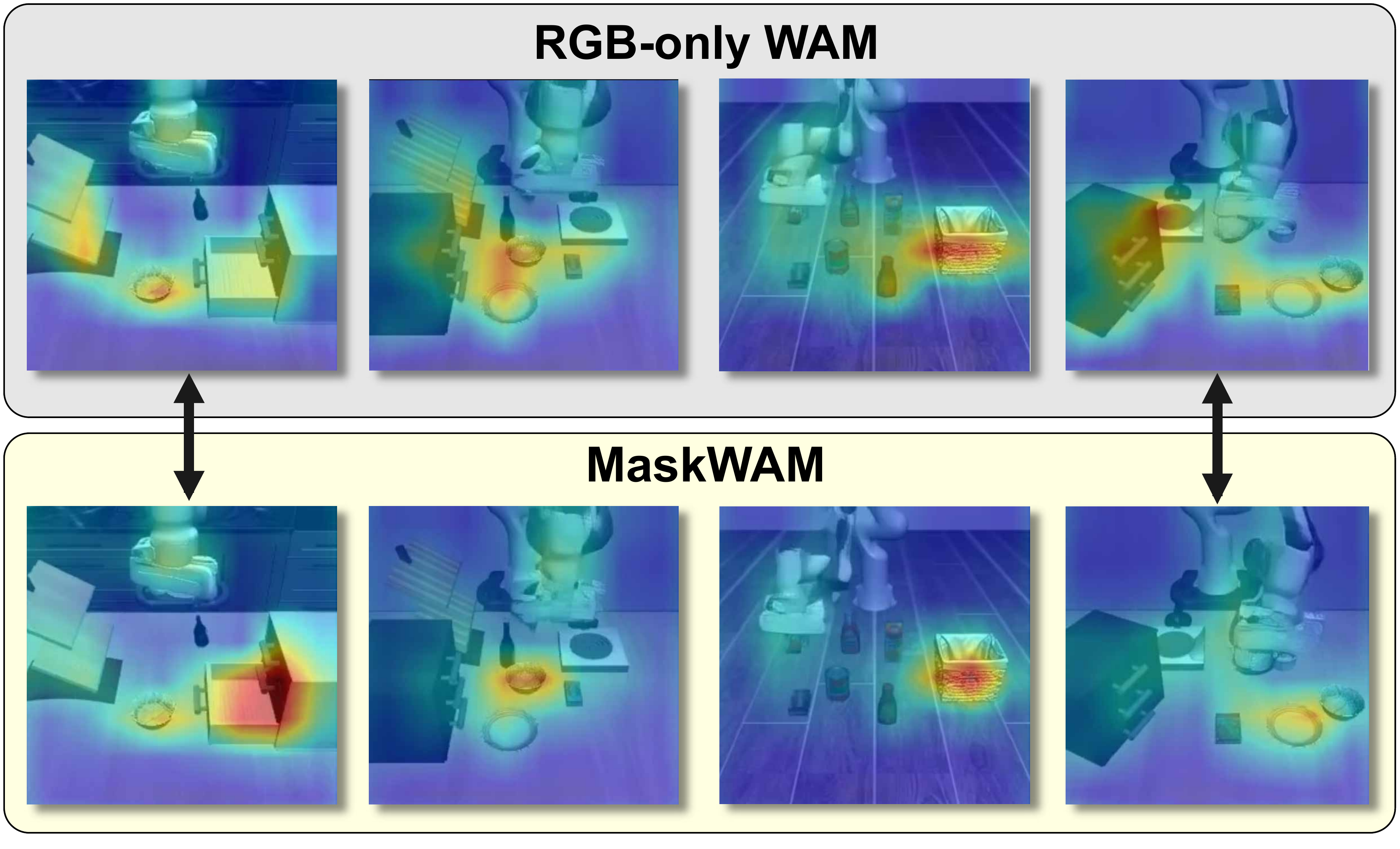}
    \vspace{-2mm}
  \end{minipage}
\end{table*}

\begin{table*}[t]
  \centering
  \footnotesize
  \setlength{\tabcolsep}{4pt} 
  \renewcommand{\arraystretch}{1.05} 
  \vspace{-2mm}
    \caption{\textbf{Success rates (\%) on the RoboTwin 2.0 benchmark.} We use 500 randomized episodes per task for training. RGB-only denotes MaskWAM with only RGB prediction, and Mask-only denotes MaskWAM with only mask prediction. }
  \vspace{-1mm}
  \label{tab:robotwin_wide}
  \begin{tabular}{lccccccc}
    \toprule
    \textbf{Method} 
    & \textbf{Hammer} 
    & \textbf{Bell} 
    & \textbf{Card} 
    & \textbf{Burger} 
    & \textbf{Stand} 
    & \textbf{Shoe} 
    & \textbf{Average} \\
    \midrule
    $\boldsymbol{\pi_0}$~\cite{black2024pi0}
    & 68 & 72 & 81 & 79 & 63 & 74 & 72.8 \\
    \textbf{FastWAM}~\cite{yuan2026fast} 
    & 83 & 87 & 92 & \cellcolor{secondgreen}\underline{94} & 80 & 90 & 87.7 \\
    \midrule
    Ours (RGB-only) 
    & 82 & 87 & 91 & 93 & 79 & \cellcolor{secondgreen}\underline{92} & 87.3 \\
    Ours (Mask-only) 
    &  \cellcolor{secondgreen}\underline{85} &  \cellcolor{secondgreen}\underline{90} &  \cellcolor{secondgreen}\underline{93} & 93 &  \cellcolor{secondgreen}\underline{81} & 91 &  \cellcolor{secondgreen}\underline{88.8}\\
    Ours 
    & \cellcolor{bestgreen}\textbf{88} & \cellcolor{bestgreen}\textbf{93} & \cellcolor{bestgreen}\textbf{95} & \cellcolor{bestgreen}\textbf{97} & \cellcolor{bestgreen}\textbf{85} & \cellcolor{bestgreen}\textbf{95} & \cellcolor{bestgreen}\textbf{92.2} \\
    \bottomrule
  \end{tabular}
  \vspace{-2mm}
\end{table*}

\subsection{Evaluation in Real-world Tasks}
\label{sec:real_world_eval}

\begin{figure}[t]
\centering
\includegraphics[width=0.9\linewidth]{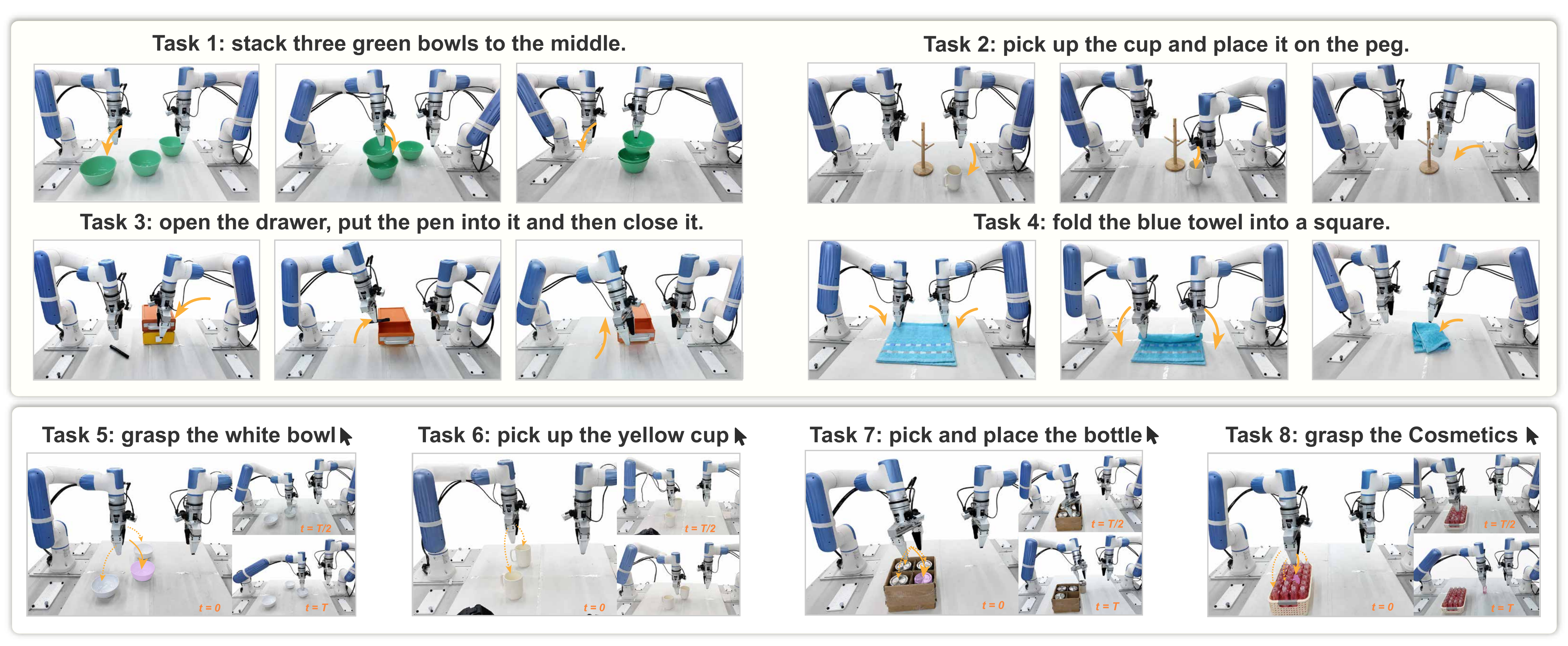}
\vspace{-1mm}
\caption{\textbf{Real-robot tasks.} Tasks 1-4 indicate standard language-clear cases, while Tasks 5-8 introduce language-ambiguous scenarios.}
\vspace{-3mm}
\label{fig:real-text-tasks}
\end{figure}

\noindent\textbf{Tasks and Evaluation Protocol.} 
To evaluate the capability and generalizability of MaskWAM, we design eight challenging real-world manipulation tasks. As illustrated in Figure~\ref{fig:real-text-tasks}, these are categorized into standard language-clear settings (Tasks 1-4) and language-ambiguous settings requiring explicit spatial prompting (Tasks 5-8). For data collection, we gather an average of 100 human demonstrations per task. For language-clear tasks, each model is evaluated over 100 trials per task. For language-ambiguous generalization, each model is evaluated over 60 trials per task under each generalization setting. Detailed descriptions for each task and their environmental setups are deferred to the Appendix.

\noindent\textbf{Performance on Language-clear Tasks.} 
To evaluate standard in-distribution performance, we first test the policies on the four fundamental text-conditioned tasks. As shown in Table~\ref{tab:text-prompt}, MaskWAM achieves highly competitive performance, surpassing state-of-the-art VLA models like $\pi_0$ and $\pi_{0.5}$, as well as prior WAM architectures such as FastWAM. Crucially, compared to our own RGB-only variant, integrating future mask prediction consistently improves success rates across the evaluated tasks, lifting performance from 86\% to 91\% on Task 1 and from 76\% to 81\% on Task 3. This indicates that jointly predicting masks does not compromise the base language-clear policy, but rather stabilizes and enhances it by providing explicit semantic grounding.

\begin{table}[htbp]
  \centering
  \footnotesize
  \setlength{\tabcolsep}{3.5pt}

  \caption{\textbf{Real-robot evaluation on language-clear tasks.} Success rates are reported in \%. We collect an average of 100 demonstrations per task, and each model is evaluated over 100 trials per task with diverse object placements.}
  \vspace{-2mm}
  \label{tab:text-prompt}
  \begin{tabular}{llccccc}
    \toprule
    \textbf{Method} & \textbf{Type} & \textbf{Task 1} & \textbf{Task 2} & \textbf{Task 3} & \textbf{Task 4} & \textbf{Avg} \\
    \midrule
    $\pi_0$~\cite{black2024pi0}             & VLA & 57 & 54 & 54 & 58 & 55.8  \\
    $\pi_{0.5}$~\cite{black2025pi0.5}       & VLA & 83 & 55 & 74 & 77 & 72.3 \\
    FastWAM~\cite{yuan2026fast}             & WAM & \cellcolor{secondgreen}\underline{88} & 76 & \cellcolor{secondgreen}\underline{77} & 75 & 79.0 \\
    \midrule
    Ours (RGB-only)                         & WAM & 86 & \cellcolor{secondgreen}\underline{77} & 76 & \cellcolor{secondgreen}\underline{78} & \cellcolor{secondgreen}\underline{79.3}  \\
    Ours                                    & WAM & \cellcolor{bestgreen}\textbf{91} & \cellcolor{bestgreen}\textbf{82} & \cellcolor{bestgreen}\textbf{81} & \cellcolor{bestgreen}\textbf{83} & \cellcolor{bestgreen}\textbf{84.3} \\
    \bottomrule
  \end{tabular}
\end{table}

\begin{figure}[htbp]
\centering
\vspace{-1mm}
\includegraphics[width=0.85\linewidth]{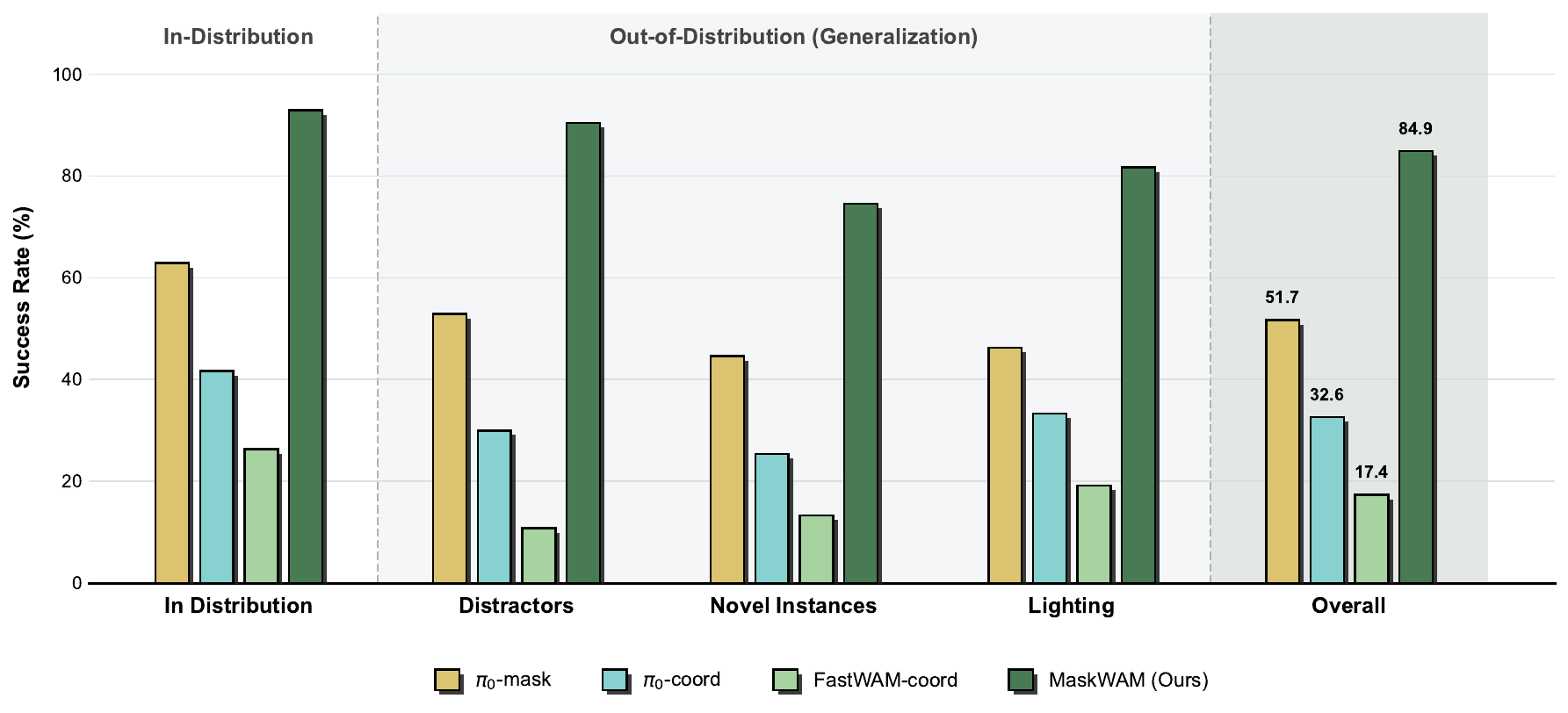}
\caption{\textbf{Generalization performance under language ambiguity with explicit spatial prompting.} We evaluate policy robustness in scenes where language instructions alone are insufficient to specify the target, demonstrating how precise spatial cues (e.g., visual masks or textual coordinate prompts) resolve target uncertainty.}
\vspace{-2mm}
\label{fig:language-ambiguity}
\end{figure}

\noindent\textbf{Performance on Language-ambiguous Tasks.} 
To evaluate how explicit spatial prompting resolves target uncertainty, we conduct a comprehensive assessment detailed in Fig.~\ref{fig:language-ambiguity}, spanning four distinct settings: one foundational in-distribution evaluation and three demanding zero-shot generalization axes. (1) \textit{In-Distribution:} Establishing foundational performance, MaskWAM achieves a 92.9\% average success rate. Notably, when utilizing textual coordinate prompts, the language-driven $\pi_0$-\texttt{coord} outperforms the vision-driven FastWAM-\texttt{coord} with a 41.7\% success rate compared to 26.3\%. This performance inversion supports our hypothesis that spatial disambiguation in WAMs is better injected natively via the visual modality of masks rather than text.

Building upon this robust visual grounding, we evaluate zero-shot generalization across the remaining three axes. (2) \textit{Distractors:} Introducing unseen, task-irrelevant objects to test spatial grounding amidst visual clutter, MaskWAM maintains a 90.4\% success rate with minimal degradation, substantially outperforming the $\pi_0$-\texttt{mask} baseline at 52.9\%. (3) \textit{Novel Instances:} Replacing the training target with unseen objects of varying geometries or related categories (e.g., substituting a trained bowl with a novel cup), MaskWAM achieves a 74.6\% success rate, significantly surpassing the 44.6\% achieved by $\pi_0$-\texttt{mask} and demonstrating robust category-level structural skill transfer. (4) \textit{Lighting:} Altering environmental illumination to test robustness against appearance shifts, MaskWAM exhibits strong invariance with an 81.7\% success rate. Across all settings, our mask-prediction mechanism consistently outperforms both visual-prompt VLA and textual coordinate baselines, confirming its efficacy in resolving language ambiguity and enhancing the visual generalization. Qualitative visualizations are provided in the appendix.

\subsection{Ablations}
\label{sec:ablations}

To validate the core architectural designs of MaskWAM, we perform controlled ablations across both simulated and real-world environments. We investigate the necessity of joint modality prediction and explicit visual prompting by addressing three critical questions:

\noindent\textbf{Q1: Is paired RGB and mask prediction essential for robust policy learning?}
Evaluated on LIBERO (Table~\ref{tab:libero}), we compare our full MaskWAM against single-objective variants. While isolated RGB-only (97.3\%) and Mask-only (97.6\%) models perform well, our joint model attains a SOTA 98.4\%. This synergy is critical for spatial precision, yielding absolute improvements of +2.0\% (\textit{Spatial}) and +1.2\% (\textit{Goal}) over the RGB-only baseline. Crucially, jointly optimizing both modalities overcomes RGB's vulnerability to visual noise and the mask's lack of textural context. By forcing the latent space to better align visual dynamics with object-centric semantics, paired prediction acts as an essential representational regularizer.

\noindent\textbf{Q2: Is future mask prediction necessary for grounding visual prompts?}
Evaluated in language-ambiguous real-world tasks (Table~\ref{tab:mask-ablation-attention}), providing a mask prompt without future mask prediction (Ours-no-pred) reduces the success rate to 21.6\%. This indicates that a visual prompt alone does not guarantee effective grounding: without the auxiliary prediction objective, the model fails to reliably use the visual guidance. In contrast, our full model achieves 84.9\%, showing that future mask prediction is essential for grounding policy attention on the visual prompt.

\noindent\textbf{Q3: How do different prompt formulations impact performance?}
Evaluated in our language-ambiguous setup (Table~\ref{tab:mask-ablation-attention}), we compare mask conditioning against spatial coordinate text (Ours-\texttt{coord}). While coordinate text struggles with spatial alignment (18.2\% success), our mask formulation (84.9\%) provides dense spatial-semantic priors to resolve referential ambiguities. As visualized in the right panel of Table~\ref{tab:mask-ablation-attention}, the text-coordinate-prompted model suffers from dispersed attention distracted by background clutter. In contrast, our mask-conditioned policy maintains a sharply localized, temporally consistent focus on the target object. Furthermore, ablation studies detailed in the Appendix confirm the model's robustness against varying input mask qualities.

\begin{table*}[htbp]
    \centering
    \footnotesize
    \setlength{\tabcolsep}{4.5pt}
    \renewcommand{\arraystretch}{1.25}
    \vspace{-2mm} 
\caption{\textbf{Ablation of mask conditioning and prediction.}
The left panel isolates first-frame mask conditioning and future mask prediction; Ours\texttt{-no-pred} only removes the future mask target and loss from MaskWAM.
The right panel compares coordinate-augmented text prompts with mask-based visual prompts.
``Coord. Prompt'' denotes coordinate-augmented textual prompting.}
    \label{tab:mask-ablation-attention}

  \begin{minipage}[t]{0.58\textwidth}
    \centering
    \vspace{0pt}
    % �� 绝对不要在这里加 \resizebox！直接用 tabular[t]
    \begin{tabular}[t]{lccc}
      \toprule
      \multicolumn{4}{c}{\textbf{Language-ambiguous Tasks (ID \& OOD)}} \\
      \midrule
      \textbf{Setting} 
      & \textbf{Ours-\texttt{no-pred}} 
      & \textbf{Ours-\texttt{coord}} 
      & \textbf{Ours} \\
      \midrule
      Future Mask 
      & -- & \checkmark & \checkmark \\
      Mask Prompt 
      & \checkmark & -- & \checkmark \\
      Coord. Prompt
      & -- & \checkmark & -- \\
      \midrule
      Avg. Succ. 
      & 21.6 & 18.2 & \textbf{84.9} \\
      \bottomrule
    \end{tabular}
  \end{minipage}
  \hfill
  % 把右边 minipage 的宽度稍微放宽一点点到 0.45，这样两边视觉更平衡
  \begin{minipage}[t]{0.4\textwidth}
    \centering
    % �� 核心对齐魔法：用负数把图片拽上来，跟左边的顶线平齐！
    \vspace{-2mm} 
    \includegraphics[width=\linewidth]{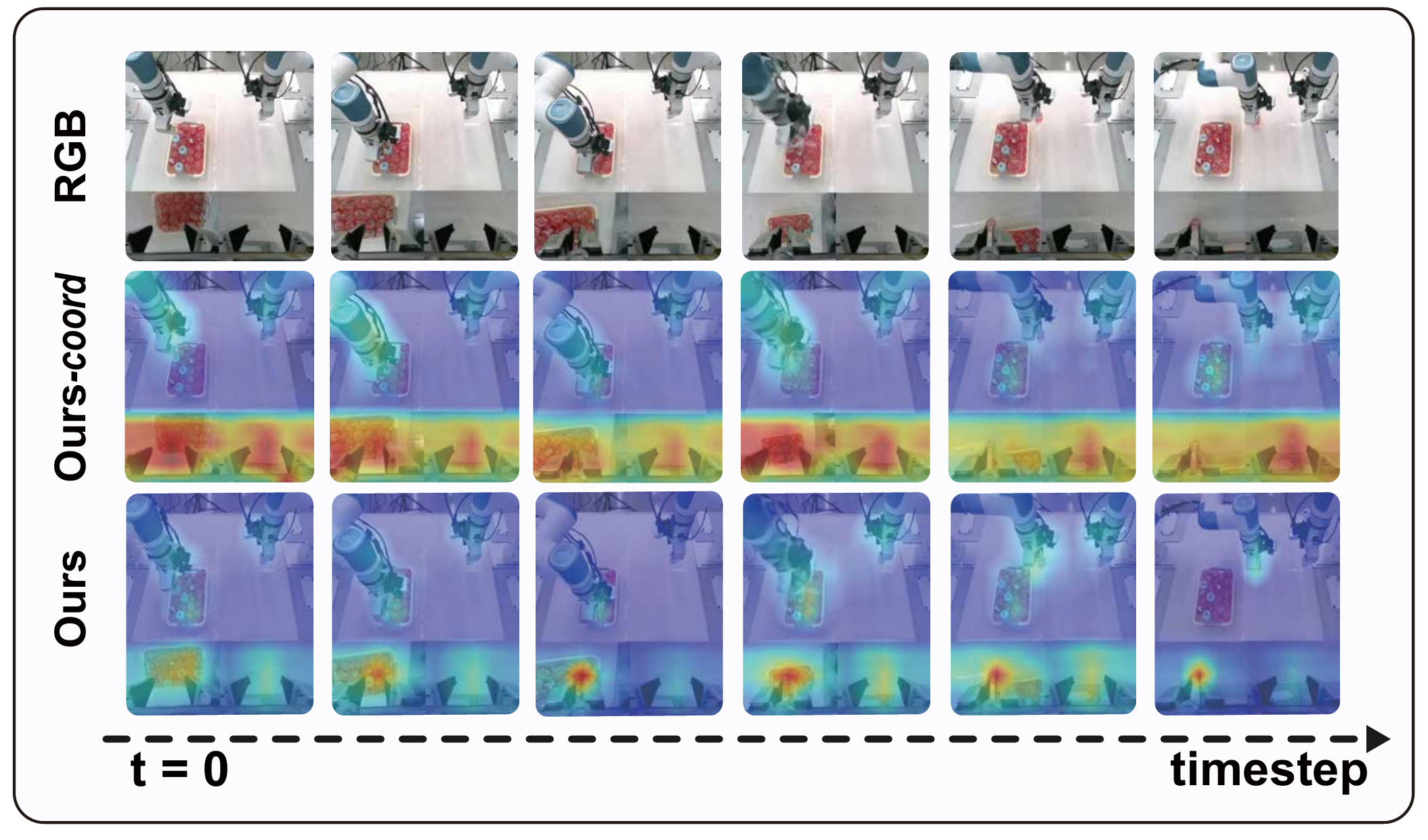}
  \end{minipage}
\end{table*}
\vspace{-2mm}

\vspace{-0.2em}
\section{Conclusion}
\vspace{-0.2em}
We introduced MaskWAM, an end-to-end world-action model that augments standard RGB prediction with auxiliary future mask prediction. By integrating mask modeling into the latent space of a unified DiT, MaskWAM encourages the policy to ground its actions in task-critical regions. This formulation improves generalization across distractors, object variations, and lighting changes, while enabling precise target disambiguation through optional first-frame mask prompts. Overall, MaskWAM shows that predicting \textit{what matters} is just as crucial as predicting \textit{what happens} for robust robotic manipulation.

\vspace{-0.2em}
\section{Limitations and Future Work}
\vspace{-0.2em}
While MaskWAM demonstrates strong performance, it has several limitations. First, MaskWAM relies on mask supervision during training and segmentation-derived prompts during deployment. Automating reliable mask extraction in cluttered real-world environments remains non-trivial. Second, due to computational constraints, we defer large-scale RGB-mask-action pretraining to future work, which holds significant potential for further enhancing mask-aware visual dynamics and downstream policy robustness.

\setlength{\bibsep}{5pt}
\bibliography{main}
\bibliographystyle{plainnat}

\newpage
\appendix
\appendix
\setcounter{section}{0}
\renewcommand{\thesection}{\Alph{section}}

% 添加总标题
\section*{Appendix} 

\section{Details about Real-world Experiments} 
\subsection{Real-world Hardware Setup.}
\begin{figure}[htbp]
\centering
\includegraphics[width=0.6\linewidth]{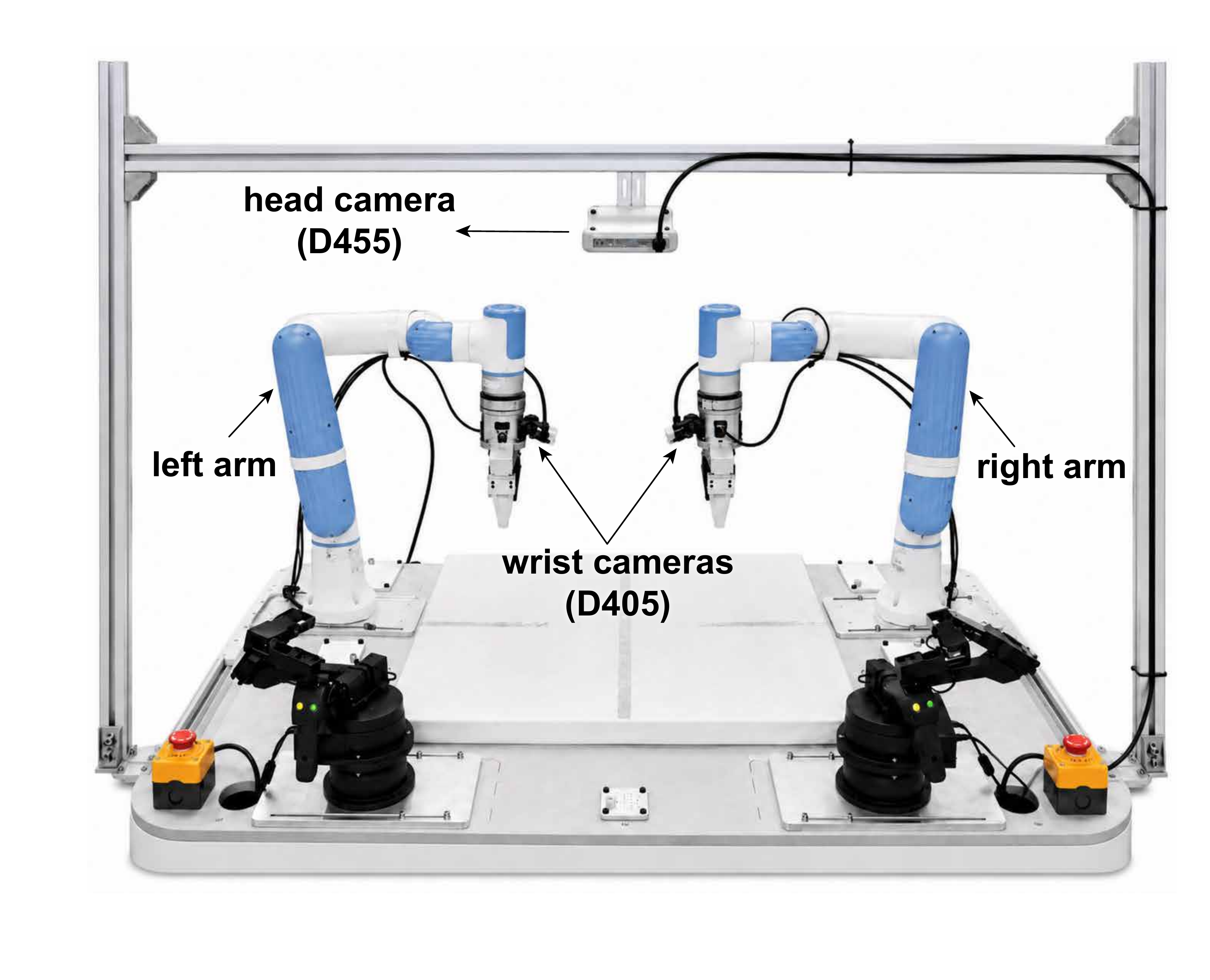}
\caption{\textbf{Real-robot platform} Our real-world experiment is based on a Dual-arm Xtrainer robotic platform}
\label{fig:xtrainer}
\end{figure}

We evaluate our model using a Dual-arm Xtrainer robotic platform, as shown in Figure \ref{fig:xtrainer}. For visual perception, we employ a multi-camera configuration: a RealSense D455 depth camera serves as the head-mounted \textit{Eye-on-Base} sensor for global scene understanding, while a RealSense D405 depth camera is integrated as an \textit{Eye-on-Hand} sensor to provide localized, high-resolution visual feedback.

\subsection{Task settings and evaluation in Real-world Tasks}
\paragraph{Task Settings.} 
To evaluate the capability and generalizability of MaskWAM, we design eight representative and challenging real-world manipulation tasks, grouped into language-clear and language-ambiguous settings as shown in Figure~\ref{fig:real-text-tasks}. The language-clear tasks include: 
(1) \textit{Stacking} three green nested bowls; 
(2) \textit{Hanging} a mug onto a designated peg on a wooden stand; 
(3) \textit{Long-horizon Interaction}, which involves opening a drawer, placing a pen inside, and closing the drawer; and 
(4) \textit{Deformable Object Manipulation}, which requires folding a fabric towel. 
The language-ambiguous tasks include: 
(5) \textit{Grasping Bowl}, grasping a specified bowl among three visually similar bowls; 
(6) \textit{Picking up Cup}, picking up the specified cup; 
(7) \textit{Picking and Placing Bottle}, picking and placing the designated bottle from a small box; and 
(8) \textit{Grasping Cosmetics}, grasping a specified cosmetic item from 16 tightly arranged red cosmetic products in a box.
We collect an average of 100 demonstrations per task. For language-clear tasks, each model is evaluated over 100 trials per task. For language-ambiguous tasks, each model is evaluated over 60 trials per task. The execution success rate is reported as the primary performance metric.

\subsection{Baseline Details for Language-Ambiguous Tasks}

In Section~\ref{sec:real_world_eval}, we compare MaskWAM with both VLA-based baselines ($\pi_0$-\texttt{mask}, $\pi_0$-\texttt{coord}) and WAM-based variants (FastWAM-\texttt{coord}). These baselines examine whether spatial information can be provided through language instructions or visual prompts. We provide the detailed baseline designs below.

For coordinate-text baselines ($\pi_0$-\texttt{coord} and FastWAM-\texttt{coord}), we first compute the centroid of the target-object mask and convert it into an explicit normalized pixel coordinate in the language instruction. For example, the command can be written as: ``grasp the white bottle whose target center is at $x=0.43$ from the left and $y=0.40$ from the top in the front view.'' We find that such coordinate-based text prompts provide a certain degree of coarse spatial control for $\pi_0$-\texttt{coord}, which achieves reasonable success rates on relatively simple tasks, such as Tasks 5 and 6, where the objects are large and well separated on the table. In contrast, FastWAM-\texttt{coord} remains largely insensitive to these coordinate-augmented instructions, even when such commands are included during training, and therefore achieves relatively low success rates. For more precision-demanding tasks, such as Tasks 7 and 8, both text-prompt baselines fail to complete the task. This suggests that even when precise coordinate information is provided in text, language conditioning remains insufficient for fine-grained manipulation. Moreover, in real-world deployment, such accurate target coordinates are usually unavailable; users can typically provide only coarse descriptions such as ``left'', ``right'', ``front'', or ``back'', which further weakens this form of spatial control.

We further investigate whether a mask image can serve as a visual prompt for guiding $\pi_0$. In the $\pi_0$-\texttt{mask} variant, the policy receives both the RGB image and the corresponding mask image. For the front view, both inputs are normalized to $[-1, 1]$ and passed through the SigLIP encoder~\cite{tschannen2025siglip} to obtain RGB and mask embeddings, respectively. To encourage the network to use the input mask, we fuse the two embeddings through direct addition:
\begin{equation}
    \mathbf{e}_{\mathrm{fused}}
    =
    \mathbf{e}_{\mathrm{rgb}}
    +
    \mathbf{e}_{\mathrm{mask}}.
\end{equation}
The fused embedding is then fed into the VLM transformer layers. We find that $\pi_0$-\texttt{mask} performs better than $\pi_0$-\texttt{coord}, but still falls short on high-precision tasks.

Overall, MaskWAM performs best, followed by $\pi_0$-\texttt{mask}, $\pi_0$-\texttt{coord}, and FastWAM-\texttt{coord}. In this comparison, WAM-based models appear less sensitive to coordinate-augmented language prompts, while MaskWAM benefits more from mask-based visual prompting. We attribute this trend to the vision-centric prediction backbone of WAMs, which allows dense visual prompts to align naturally with the visual prediction process.

% \subsection{Inference Latency}
% MaskWAM achieves an inference latency of 225 ms on an RTX 5090D using one-step partial denoising. This represents a 10$\times$ speedup compared to the full denoising strategy (2360 ms), validating the efficiency of our partial denoising design. This masks MaskWAM a practical choice for real-world deployment.

\section{Dataset Annotation and Mask-Prompt Robustness}
\label{sec:annotation_robustness}

\subsection{Language and Mask Annotation}
\label{sec:annotation}
\begin{figure}[t]
\centering
\includegraphics[width=0.95\linewidth]{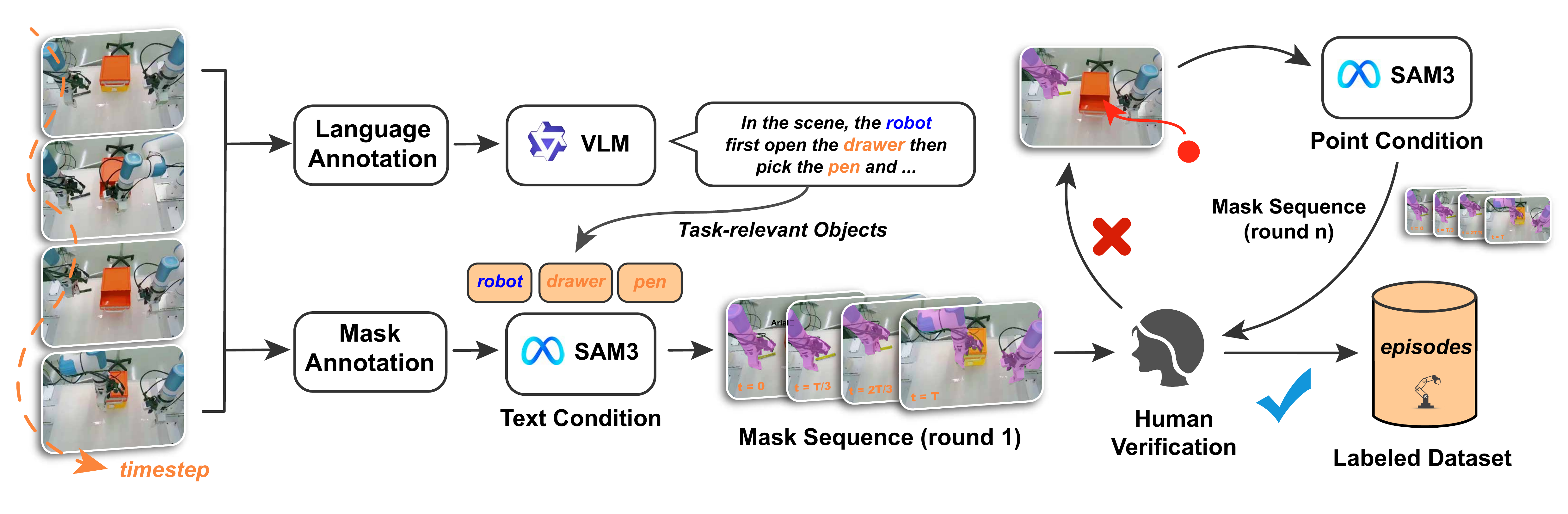}
\vspace{-0.5em}
\caption{\textbf{Annotation pipeline.} We obtain language labels, initialize SAM3 from task-relevant prompts, and propagate masks with human verification.}
\label{fig:annotation-pipeline}
\vspace{-0.8em}
\end{figure}

We build an automated annotation pipeline as shown in Figure~\ref{fig:annotation-pipeline}. The pipeline differs between language-clear and language-ambiguous tasks to avoid oracle leakage.

\textbf{Language-clear tasks:} Qwen3-VL~\cite{qwen3vl2025} parses the instruction to identify task-relevant objects, and SAM3~\cite{carion2025sam} segments and tracks masks across the episode.

\textbf{Language-ambiguous tasks:} Since text alone cannot disambiguate the target, a human annotator provides a point prompt on the first frame. SAM3 propagates this point into a full mask and tracks it across the episode.

All first-round annotations are human-verified. Low-quality episodes are corrected with point prompts on keyframes, and SAM3 re-propagates the corrected masks. This loop repeats until quality is satisfactory.
With this pipeline, 91\% of episodes require no human correction, and annotating 50 episodes takes approximately 3 minutes. For language-ambiguous tasks, the initial point prompt adds marginal overhead (5--10 seconds per episode).

\subsection{Robustness to Noisy First-frame Mask Prompts}
\label{sec:mask_prompt_robustness}

In our real-world experiments, the first-frame prompt mask is generated online by SAM3~\cite{carion2025sam} at the beginning of each episode. 
To evaluate whether MaskWAM relies on perfectly clean mask prompts, we further study its sensitivity to imperfect first-frame masks. 
We keep the trained MaskWAM checkpoint fixed and perturb only the first-frame prompt mask $M_0$ during inference. 
All subsequent observations, language instructions, proprioceptive states, and model weights remain unchanged.

\begin{figure}[t]
\centering
\includegraphics[width=1.0\linewidth]{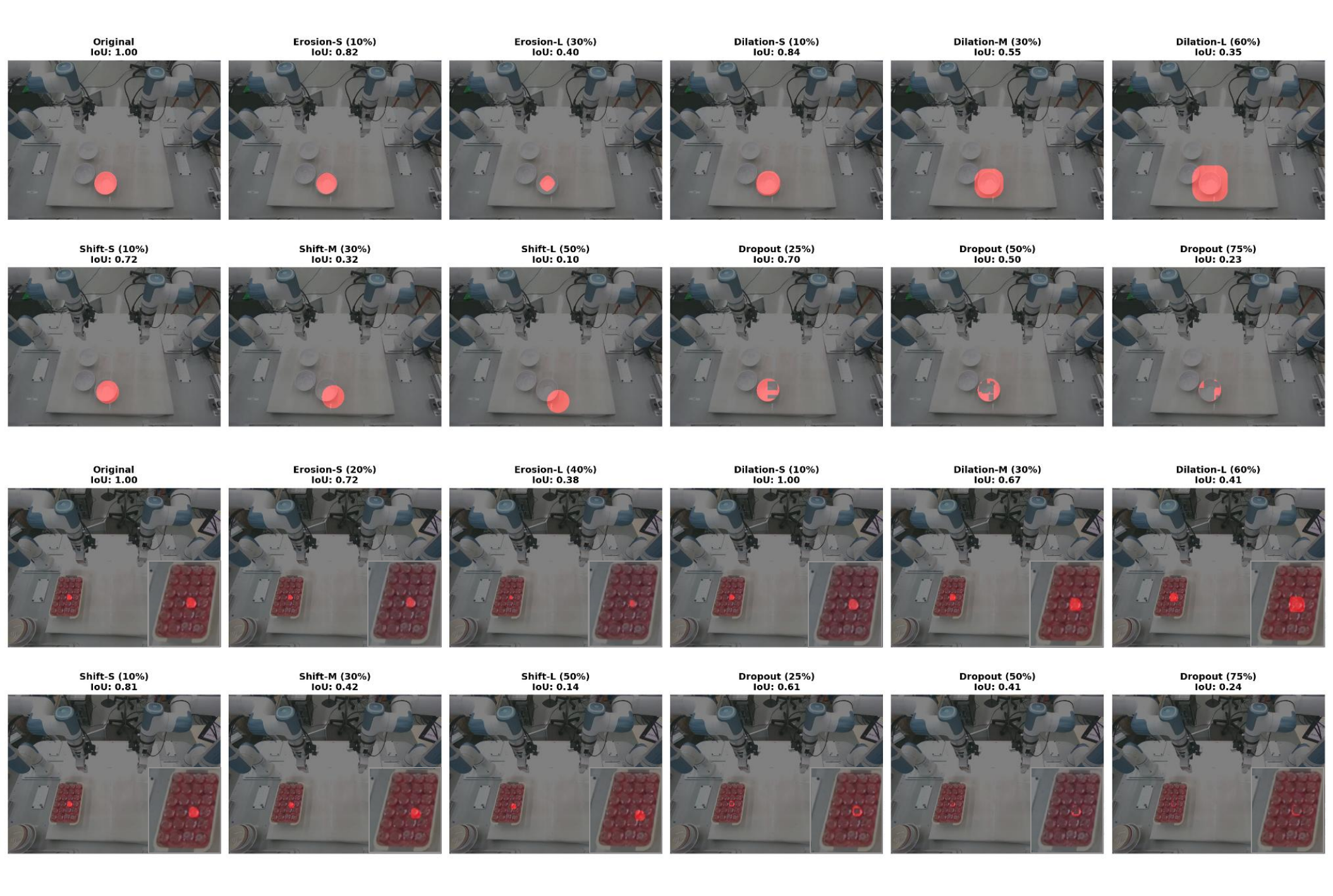}
\vspace{-0.5em}
\caption{\textbf{Examples of imperfect first-frame mask prompts.} 
We visualize representative corrupted masks generated from the online SAM3 mask, including erosion, dilation, spatial shift, and region dropout.}
\label{fig:mask_examples}
\vspace{-0.8em}
\end{figure}

We evaluate two representative real-world language-ambiguous tasks: Task 5, where the target object is relatively large and well separated, and Task 8, where the target object must be selected from densely arranged visually similar cosmetic items. 
For each task and each mask condition, we run 20 trials with diverse initial object placements. 
The default condition uses the online SAM3 mask, which is the same setting as our main real-world evaluation. 
We then apply four types of synthetic corruption to this online mask, as illustrated in Fig.~\ref{fig:mask_examples}. 
\textbf{Erosion} simulates under-segmentation, \textbf{dilation} simulates over-segmentation, \textbf{shift} simulates spatial mislocalization caused by inaccurate prompting or segmentation, and \textbf{region dropout} simulates partial mask missing caused by occlusion, reflection, or segmentation failure.

\begin{figure}[t]
\centering
\includegraphics[width=0.9\linewidth]{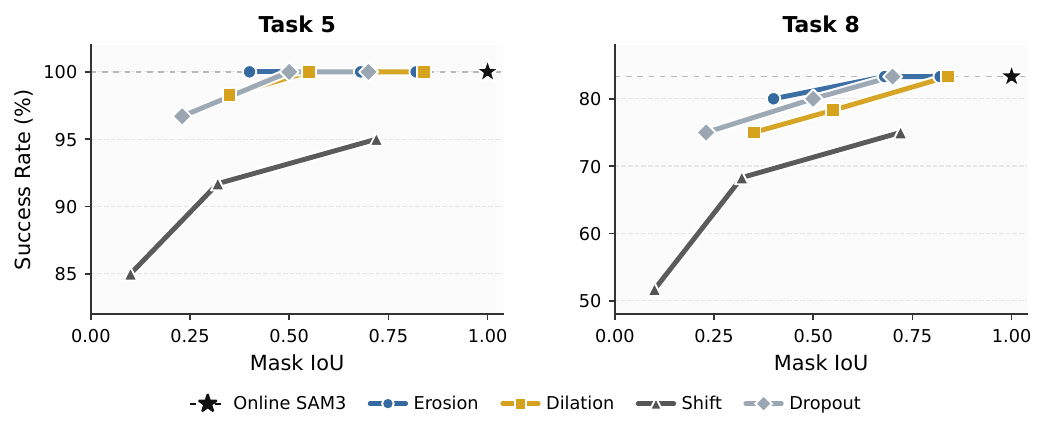}
\vspace{-0.5em}
\caption{\textbf{Robustness to noisy first-frame mask prompts.} 
We perturb only the first-frame prompt mask at inference time while keeping the model checkpoint fixed. 
The online SAM3 mask is used as the default prompt, and all corrupted masks are generated from this default mask. 
The x-axis denotes the mask IoU with respect to the online SAM3 mask before perturbation, and the y-axis denotes the real-world success rate.}
\label{fig:mask_robustness}
\vspace{-0.8em}
\end{figure}

As shown in Fig.~\ref{fig:mask_robustness}, MaskWAM is robust to moderate first-frame prompt corruption. 
Starting from the online SAM3 mask used in our main real-world evaluation, mild erosion, dilation, shift, and region dropout only lead to limited performance degradation, especially when the corrupted mask remains sufficiently overlapped with the original target mask. 
This suggests that MaskWAM does not require a perfectly clean prompt mask, as long as the mask still provides an approximate target identity and spatial anchor. 
In contrast, performance drops more clearly under severe corruption, such as large spatial shifts or heavy region dropout, where the prompt either becomes mislocalized or loses most of the target area. 
These results indicate that the first-frame mask primarily serves as a target identity and spatial anchoring signal: MaskWAM tolerates moderate mask noise, but severe target-region corruption weakens object-centric grounding.

\begin{table}[t]
\centering
\footnotesize
\caption{\textbf{Generalization performance under language ambiguity with explicit spatial prompting.} 
Success rates are reported in \%, and each model is evaluated over 60 trials per task under each generalization setting. 
Additional qualitative examples of these generalization settings are shown in Figure.~\ref{fig:moredata}.}
\label{tab:generalization_mask}
\setlength{\tabcolsep}{3.8pt}
\renewcommand{\arraystretch}{1.05}
\begin{tabular}{lllccccc}
\toprule
\multirow{2}{*}{\makecell[l]{Generalization\\Setting}} & \multirow{2}{*}{Method} & \multirow{2}{*}{Type} & \multicolumn{4}{c}{Language ambiguity} & \multirow{2}{*}{\makecell{Average\\(\%)}} \\
\cmidrule(lr){4-7}
& & & Task 5 & Task 6 & Task 7 & Task 8 & \\
\midrule

\multirow{4}{*}{In Domain} 
& $\pi_0$-\texttt{mask} & VLA & 96.7 & 93.3 & 38.3 & 23.3 & 62.9 \\
& $\pi_0$-\texttt{coord} & VLA & 53.3 & 88.3 & 20.0 & 5.0 & 41.7 \\
& FastWAM-\texttt{coord} & WAM & 35.0 & 55.0 & 13.3 & 1.7 & 26.3 \\
& \textbf{Ours} & WAM & \textbf{100.0} & \textbf{100.0} & \textbf{88.3} & \textbf{83.3} & \textbf{92.9} \\

\midrule

\multirow{4}{*}{Distractors} 
& $\pi_0$-\texttt{mask} & VLA & 86.7 & 75.0 & 31.7 & 18.3 & 52.9 \\
& $\pi_0$-\texttt{coord} & VLA & 46.7 & 61.7 & 11.7 & 0.0 & 30.0 \\
& FastWAM-\texttt{coord} & WAM & 15.0 & 23.3 & 5.0 & 0.0 & 10.8 \\
& \textbf{Ours} & WAM & \textbf{100.0} & \textbf{98.3} & \textbf{85.0} & \textbf{78.3} & \textbf{90.4} \\

\midrule

\multirow{4}{*}{Novel Instances} 
& $\pi_0$-\texttt{mask} & VLA & 71.7 & 65.0 & 26.7 & 15.0 & 44.6 \\
& $\pi_0$-\texttt{coord} & VLA & 41.7 & 51.7 & 8.3 & 0.0 & 25.4 \\
& FastWAM-\texttt{coord} & WAM & 18.3 & 28.3 & 6.7 & 0.0 & 13.3 \\
& \textbf{Ours} & WAM & \textbf{86.7} & \textbf{76.7} & \textbf{70.0} & \textbf{65.0} & \textbf{74.6} \\

\midrule

\multirow{4}{*}{Lighting} 
& $\pi_0$-\texttt{mask} & VLA & 75.0 & 68.3 & 28.3 & 13.3 & 46.3 \\
& $\pi_0$-\texttt{coord} & VLA & 48.3 & 66.7 & 15.0 & 3.3 & 33.3 \\
& FastWAM-\texttt{coord} & WAM & 31.7 & 36.7 & 8.3 & 0.0 & 19.2 \\
& \textbf{Ours} & WAM & \textbf{93.3} & \textbf{91.7} & \textbf{73.3} & \textbf{68.3} & \textbf{81.7} \\
\bottomrule
\end{tabular}
\vspace{1mm}
\end{table}

\section{Ablations on Alternative Mask-Conditioning Designs}
\label{app:things_not_work}

Here we discuss several design choices we considered before settling on the final MaskWAM formulation described in Section~\ref{sec:method}. These negative results suggest that masks should not be treated as a lightweight side signal, but should be aligned with the pretrained visual latent space and coupled with an explicit future prediction objective.

\paragraph{Direct mask downsampling or a separate mask encoder.}
We first experimented with simpler ways of obtaining mask latents. One option was to directly downsample binary masks to the spatial resolution of the video latent. Another option was to train a lightweight 3D CNN mask encoder from scratch. Both designs led to worse performance than encoding rendered mask frames with the pretrained RGB video VAE. We found that the choice of VAE is important: directly downsampled masks lack the semantic and structural priors of the pretrained visual latent space, while a separately learned mask encoder introduces a representation mismatch between RGB latents and mask latents. In contrast, using the same RGB VAE for rendered masks maps both RGB and mask observations into a shared latent space, making channel-level fusion more stable.

\paragraph{ControlNet-style mask injection.}
We also tried injecting mask information through a separate mask encoder and zero-initialized residual adapters, similar in spirit to ControlNet-style conditioning. This design did not work well in our setting. Although the side branch provides an additional mask condition, the mask signal remains auxiliary and is not explicitly tied to the model's future prediction objective. As a result, the policy can still underuse the mask, especially in language-ambiguous tasks where the target object must be distinguished from visually similar distractors. This motivated us to make the mask part of the main RGB-mask latent stream and supervise it directly through future mask prediction.

\paragraph{Zero-initialized gated mask fusion for VLA.}
For the visual-prompted VLA baseline, we further tested a zero-initialized gated fusion design. Specifically, the RGB image and mask image are encoded separately, and the fused visual embedding is computed as
\[
e_{\mathrm{fused}} = e_{\mathrm{rgb}} + g \cdot e_{\mathrm{mask}},
\]
where the learnable gate $g$ is initialized to zero to preserve the original pretrained behavior at the beginning of fine-tuning. However, we observed that the model tended to keep the mask pathway weak and largely ignored the mask embedding during training. This suggests that preserving the pretrained behavior through zero initialization, while useful for stability, does not by itself guarantee that the model will learn to use the new mask signal for fine-grained spatial grounding.

\begin{figure}[t]
\centering
\includegraphics[width=1.0\linewidth]{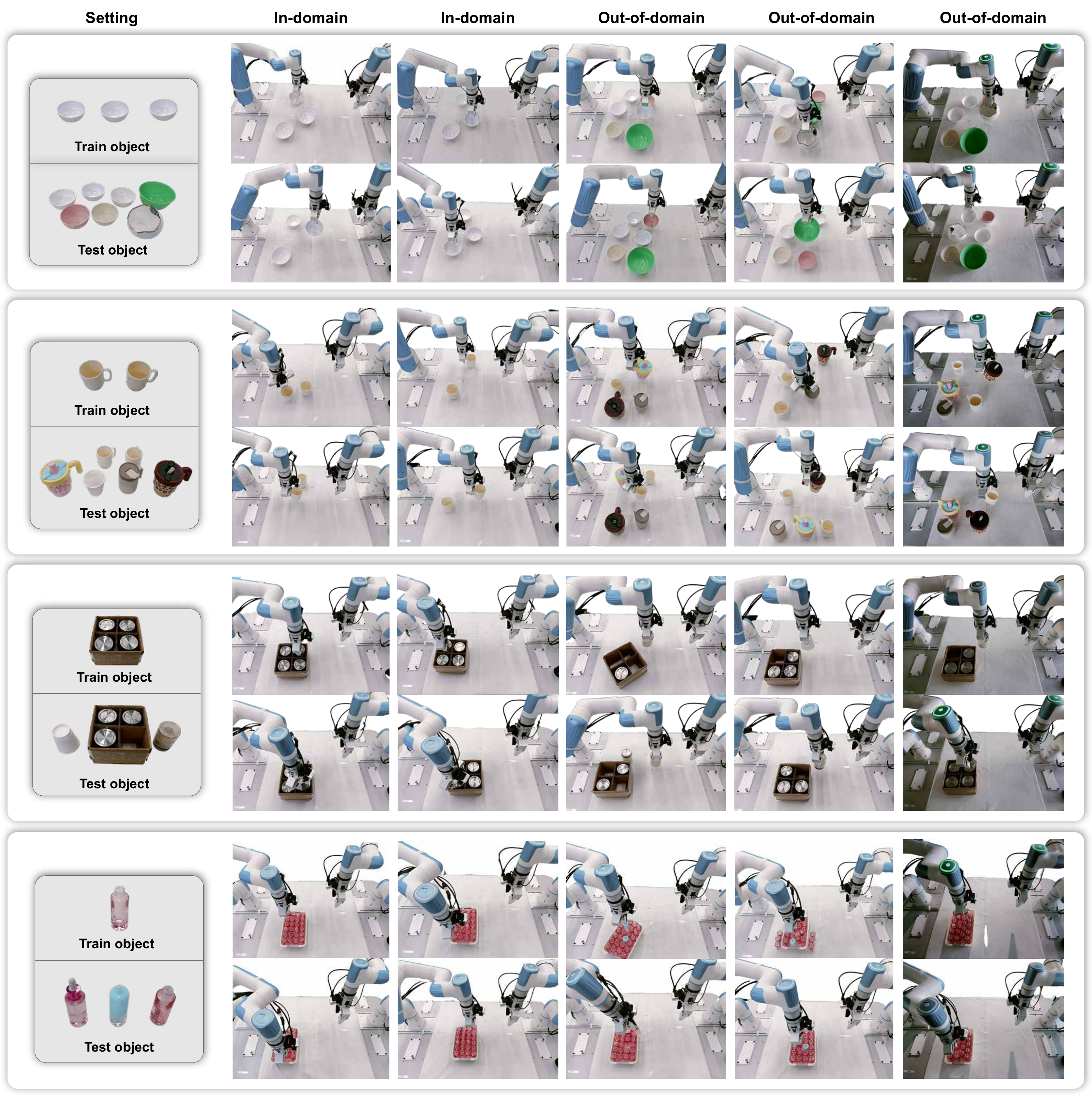}
\caption{\textbf{Additional in-domain and out-of-domain real-world demonstrations.} We show representative in-domain and out-of-domain rollouts, demonstrating that MaskWAM maintains precise mask-based target grounding under varied object appearances, placements, and scene layouts.} 
\label{fig:moredata}
\end{figure}

\begin{figure}[t]
\centering
\includegraphics[width=0.9\linewidth]{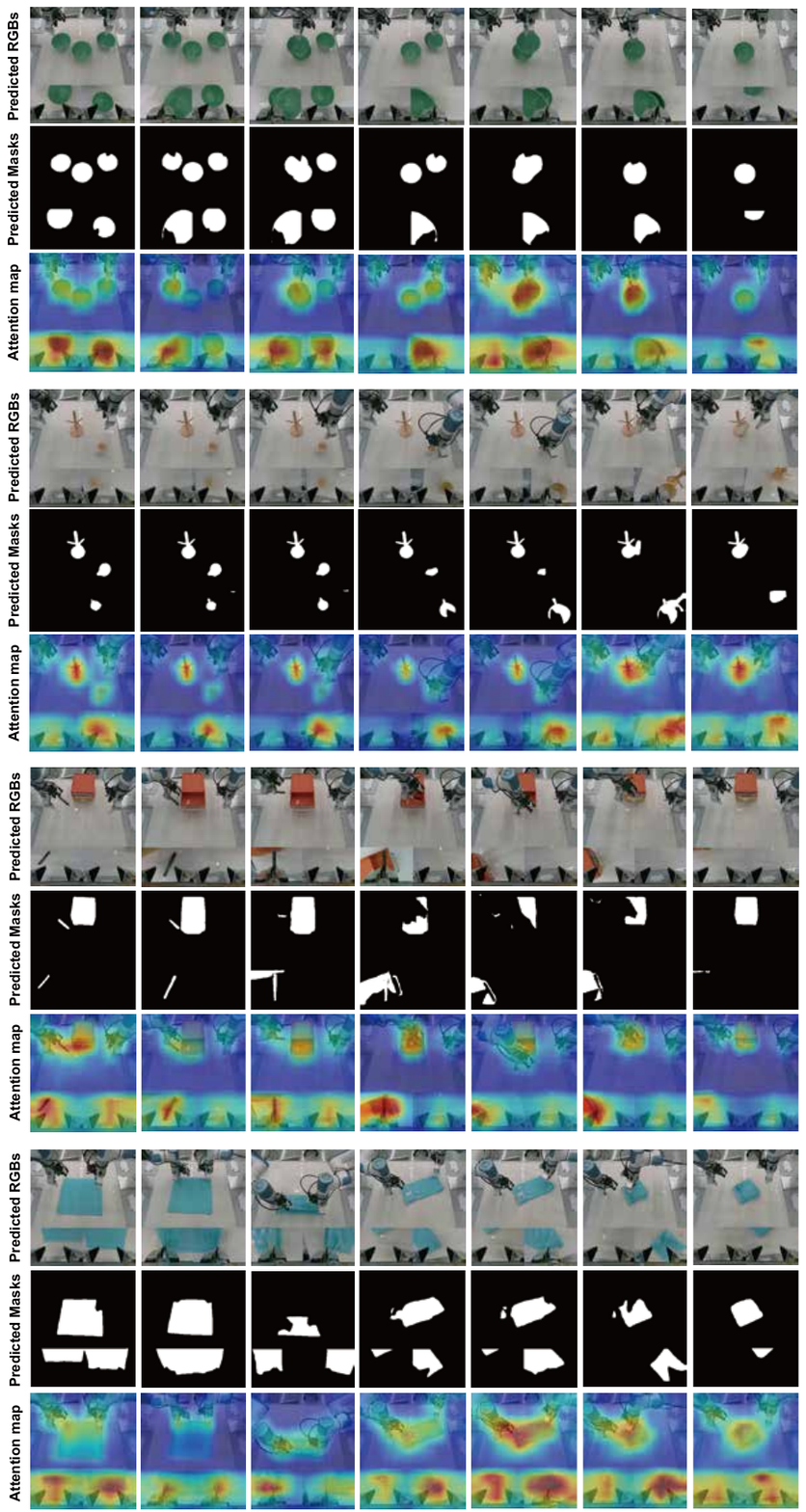}
\caption{\textbf{Visualization of predicted RGB frames, masks, and attention masks in real data.}
For visualization only, we decode full future RGB and mask sequences offline; during real-world deployment, MaskWAM uses partial denoising and does not generate full future sequences at test time.}
\label{fig:attn_more}
\end{figure}

\begin{figure}[t]
\centering
\includegraphics[width=0.8\linewidth]{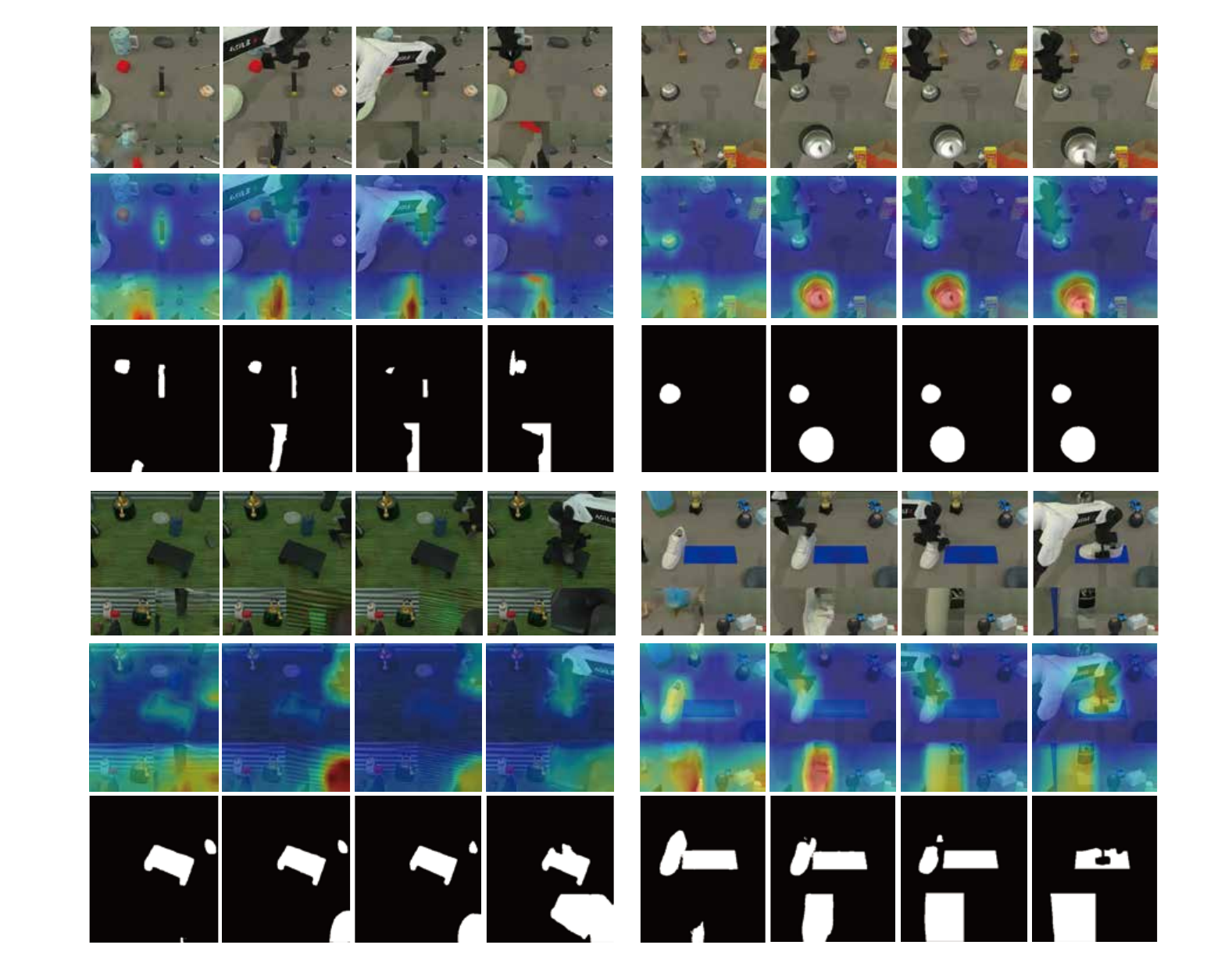}
\caption{\textbf{Visualization of predicted RGB frames, masks, and attention masks in Robotwin benchmark.}}
\label{fig:attn_more}
\end{figure}

\begin{figure}[t]
\centering
\includegraphics[width=0.8\linewidth]{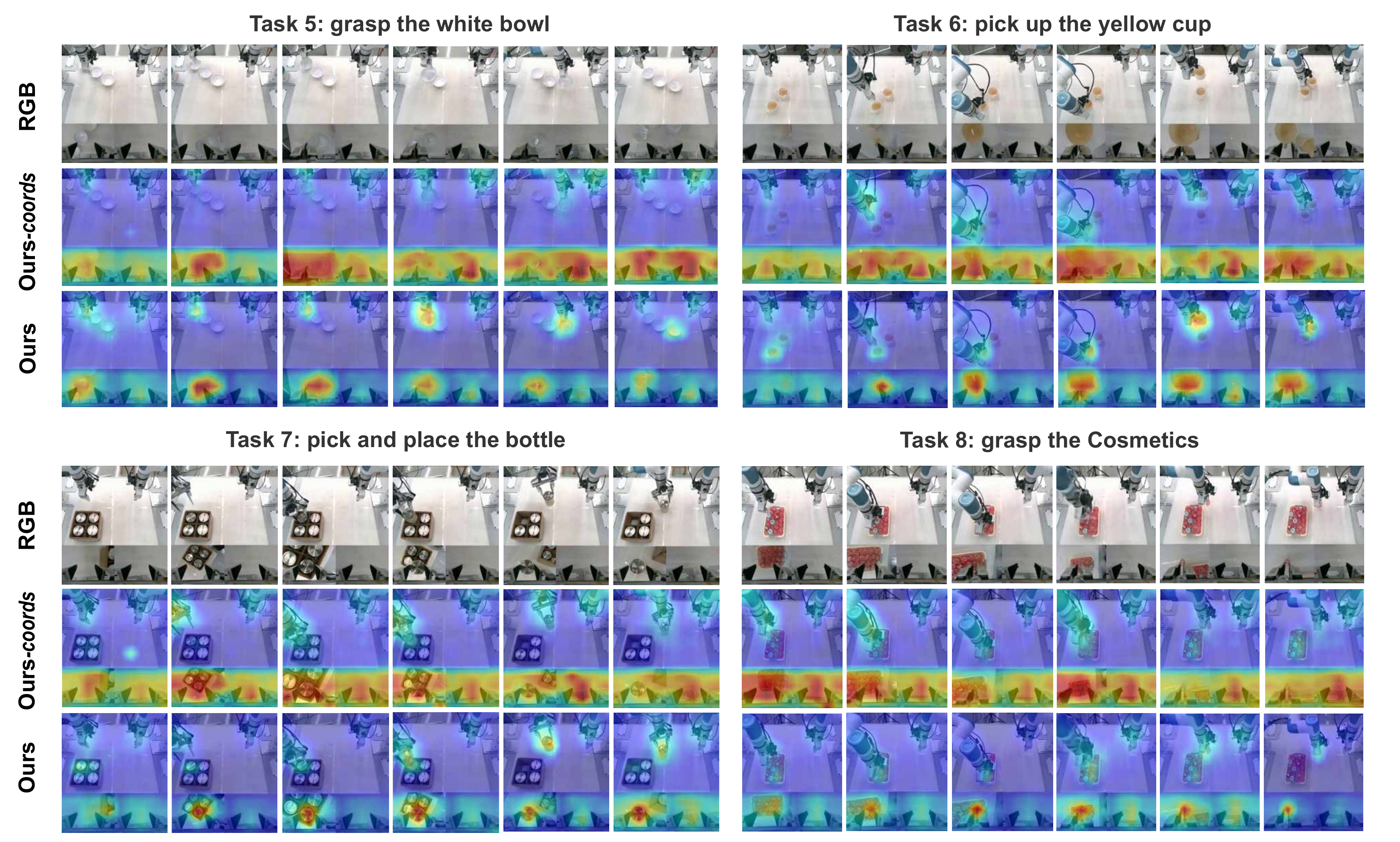}
\caption{\textbf{Comparison of attention maps between text-prompted and visual-prompted WAMs.}
The text-prompted WAM fails to reliably distinguish different textual instructions, leading to scattered attention patterns over task-irrelevant regions. In contrast, with the first-frame mask prompt and future mask prediction, MaskWAM produces more focused attention on task-relevant regions.}
\label{fig:attn_more}
\end{figure}

\end{document}